\documentclass{article} 
\usepackage{iclr2025_delta,times}

\usepackage{amsmath,amsfonts,bm}

\def\eqref#1{equation~\ref{#1}}

\def\floor#1{\lfloor #1 \rfloor}
\def\1{\bm{1}}

\DeclareMathAlphabet{\mathsfit}{\encodingdefault}{\sfdefault}{m}{sl}
\SetMathAlphabet{\mathsfit}{bold}{\encodingdefault}{\sfdefault}{bx}{n}

\usepackage{hyperref}
\usepackage{url}
\usepackage{microtype}
\usepackage{graphicx}
\usepackage{subcaption}
\usepackage{enumitem}
\usepackage{booktabs} 
\usepackage{pifont}
\usepackage{mathtools}
\usepackage[capitalise]{cleveref}  
\usepackage{wrapfig} 
\usepackage{caption}
\usepackage[most]{tcolorbox} 
\usepackage{multirow}

\title{Designing Parameter and Compute Efficient \newline Diffusion Transformers using Distillation}

\author{Vignesh Sundaresha \\
University of Illinois Urbana Champaign \\
\texttt{vs49@illinois.edu}
\vspace{-10pt}
}

\iclrfinalcopy 
\begin{document}

\maketitle

\begin{abstract}
Diffusion Transformers (DiTs) with billions of model parameters form the backbone of popular image and video generation models like DALL.E, Stable-Diffusion and SORA. Though these models are necessary in many low-latency applications like Augmented/Virtual Reality, they cannot be deployed on resource-constrained Edge devices (like Apple Vision Pro or Meta Ray-Ban glasses) due to their huge computational complexity. To overcome this, we turn to knowledge distillation and perform a thorough design-space exploration to achieve the best DiT for a given parameter size. In particular, we provide principles for how to choose design knobs such as depth, width, attention heads and distillation setup for a DiT. During the process, a three-way trade-off emerges between model performance, size and speed that is crucial for Edge implementation of diffusion. We also propose two distillation approaches - Teaching Assistant (TA) method and Multi-In-One (MI1) method - to perform feature distillation in the DiT context. Unlike existing solutions, we demonstrate and benchmark the efficacy of our approaches on practical Edge devices such as NVIDIA Jetson Orin Nano.
\end{abstract}
\vspace{-15pt}
\section{Introduction}
\label{sec: intro}
\vspace{-5pt}
Diffusion Transformers (DiTs)~\citep{peebles2023scalable} have become the de facto method~\citep{dhariwal2021diffusion} for generating images and videos due to their high fidelity~\citep{ho2020denoising}, generalizability~\citep{nichol2021improved}, ease of training~\citep{ho2020denoising} and scalability~\citep{peebles2023scalable}. DiTs form the backbone of various practically deployed image and video generation models like DALL.E~\citep{betker2023improving}, StableDiffusion~\citep{esser2024scaling} and Sora~\citep{sora2024}. Due to the large parameter size and computational complexity of these models, one has to employ Cloud services to run them remotely. The significant latencies associated with such data transmissions from Cloud to the Edge cannot be afforded for high-frame-rate applications like Augmented/Virtual Reality (AR/VR) which need to be implemented on resource-constrained Edge devices~\citep{apple_vision_pro, meta_orion}.

The main challenge in directly implementing neural network inferences on Edge devices comes from the limited memory and energy capacity of the Edge hardware. To address this, we need to \textit{design parameter- and compute-efficient DiTs}. Edge devices which typically hold on-chip memories in the range of few megabytes, thus requiring model sizes to be in the order of a million parameters as compared to existing practical models~\citep{peebles2023scalable, crowson2024scalable} which have billions of parameters. Prior works (explained in detail in~\cref{app: related works}) which focus on efficient DiTs optimize only specific layers~\citep{pu2024efficient} or look at only precision~\citep{wu2024ptq4dit} or do not push the parameter-limit required to achieve the desired performance~\citep{geng2024one}. 
The focus of our work is \textit{not} about providing SOTA DiT models through novel algorithmic methods. Instead, \textit{our goal is to provide the best DiT model - in terms of performance and speed - at a given parameter size using principled design choices.}

\textbf{Contributions: }
\vspace{-10pt}
\begin{enumerate}[noitemsep]
    \item We provide principles for designing an efficient SOTA (at the given model size) DiT model (DiT-Nano) by employing distillation. Through the process, we highlight a key trade-off that emerges between model performance (FID), size (\#parameters) and speed (latency). In particular, we show the practical impact of our designed models on real-life Edge devices such as NVIDIA Jetson Orin Nano. This is our key contribution.
    \item We propose two algorithms - Teaching Assistant (TA) method and the Multi-In-One (MI1) method, and explore the efficacy of these methods.
\end{enumerate}

\vspace{-10pt}

\section{Designing Efficient DiTs}
\label{sec: Design description}
\vspace{-15pt}
\begin{wrapfigure}{L}{0.6\columnwidth}
        \vskip 0.2in
        \begin{center}
            \centerline{\includegraphics[width=0.6\columnwidth]{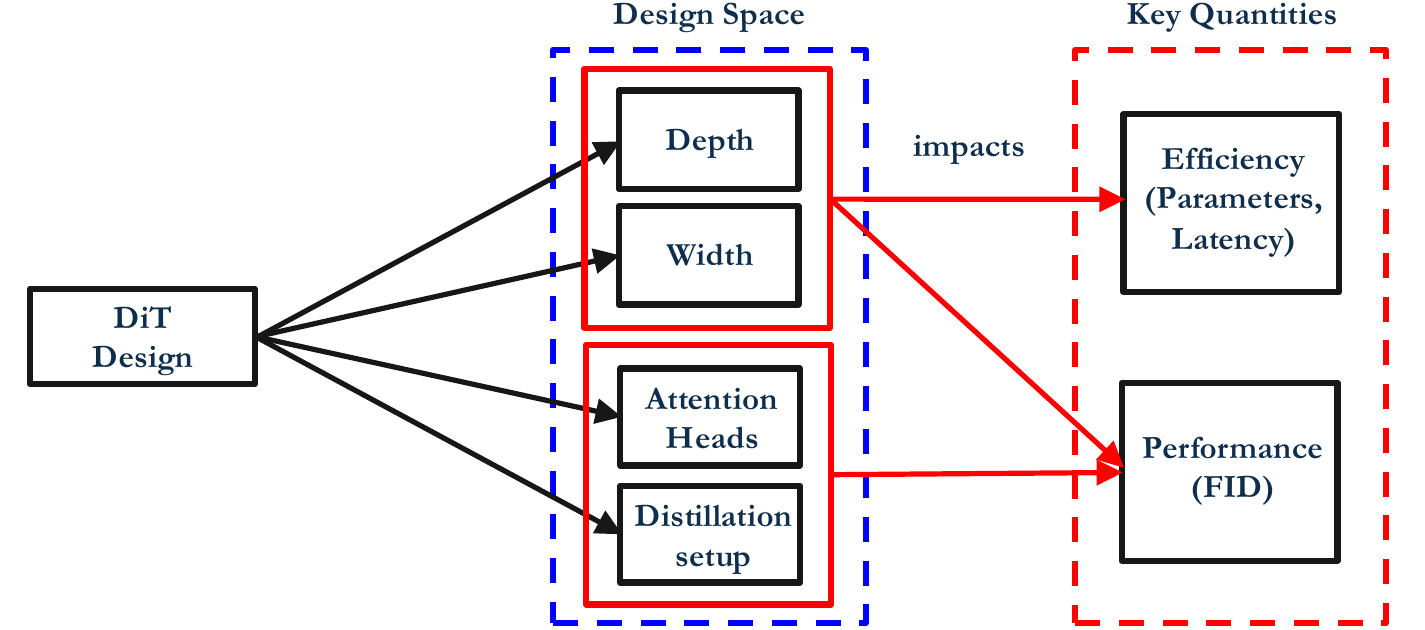}}
            \caption{Design space exploration of diffusion distillation.}
            \label{fig: design space exploration}
        \end{center}
        \vskip -0.2in
\end{wrapfigure}
\textbf{Design-Space Exploration}: Among the several design knobs for distilling DiT models, we pick the following most relevant ones - depth, width, number of attention heads of the DiT model, and the setup (loss function and teacher models) for distillation. The former two knobs impact both the efficiency and performance, while the latter two only impact performance. We do not consider timesteps as a design knob despite its importance since extensive studies have been performed on that already~\citep{peebles2023scalable, yin2024improved} (we do 1-step diffusion only). Below we propose two methods which explore new Distillation Setups for DiT.

\begin{figure}[ht]
\vskip 0.2in
\begin{center}
\centerline{\includegraphics[width=\columnwidth]{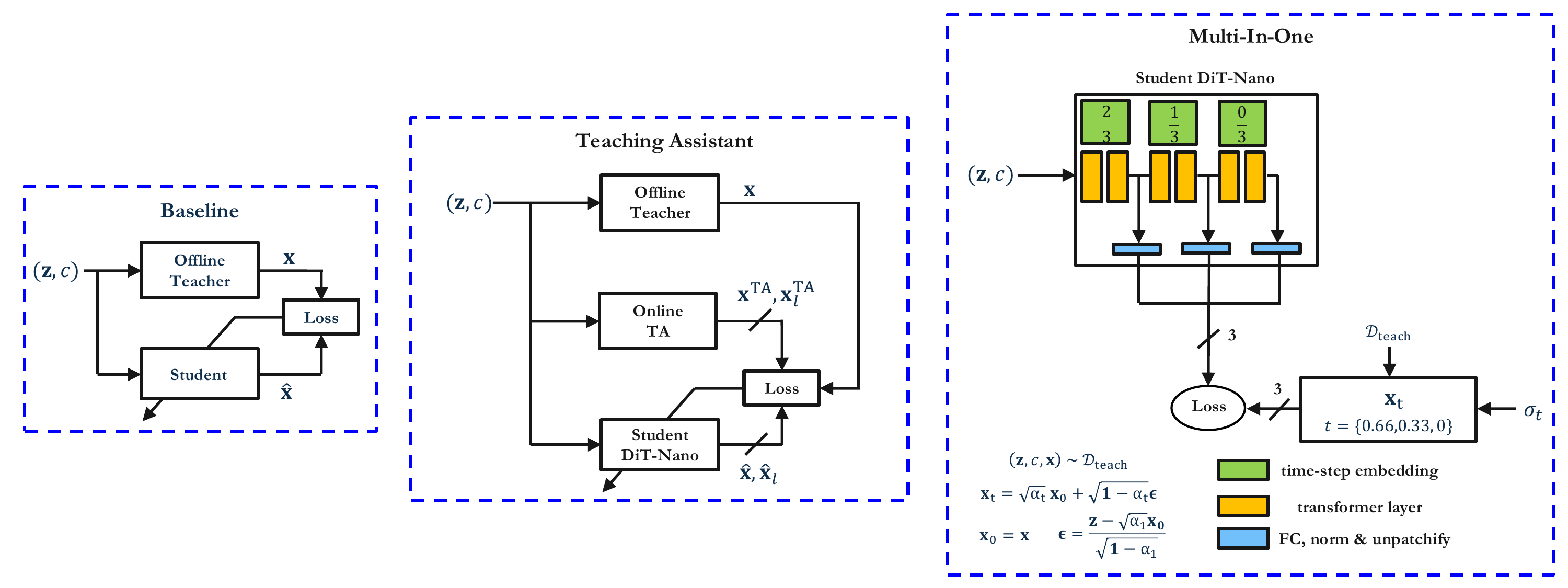}}
\caption{(Left) Baseline approach which performs regular knowledge distillation using offline teacher~\citep{geng2024one}. (Center) Teaching Assistant (TA) which performs layer-wise distillation using an online teaching assistant and an offline teacher. (Right) Multi-In-One (MI1) approach which maps multiple diffusion steps into a single step. This is done by mapping different diffusion timesteps to different layers of a DiT. The training targets are obtained using forward diffusion of the probability flow ODE. See~\cref{app: algorithms} for implementation details of TA and MI1.}
\label{fig: distillation methodology}
\end{center}
\vskip -0.2in
\end{figure}

\textbf{Teaching Assistant (TA) Method}: This approach is inspired by the original TA paper~\citep{mirzadeh2020improved} for distilling convolutional networks. We explore the possibility of combined feature distillation (see~\cref{fig: distillation methodology}) using the teacher and TA with LPIPS loss~\citep{zhang2018unreasonable}. 

\textbf{Multi-In-One (MI1) Method}: This approach performs multiple diffusion timesteps in a single step by mapping the diffusion samples to specific layers of the student. The noise-image pair of the teacher model which does multi-step diffusion is used to calculate the intermediate noisy images by employing the forward diffusion probability flow ODE (shown in~\cref{appsubsec: MI1}). 

\vspace{-10pt}
\vspace{-5pt}
\vspace{-10pt}
\section{Experiments}
\vspace{-5pt}
In this section we mainly describe the principles for designing efficient DiTs and the experiments to corroborate them. The figure captions are provided with extensive detail and hence the text will only include points not mentioned in the captions.
\vspace{-5pt}
\subsection{Setup}
\vspace{-5pt}
The comprehensive details of the training setup is provided in the~\cref{app: setup}. The results below are shown for CIFAR-10 on DiTs distilled using EDM~\citep{karras2022elucidating} as teacher. We do not show results for other larger models or datasets since such a design-space exploration is computationally infeasible. Thus we provide guidelines from our study for such bigger and more practical implementations. We use FID~\citep{heusel2017gans} as the metric to evaluate the generated image performance while using model size and latency (instead of FLOPs) as the metrics for efficiency. Even though our results can be extended to other scenarios, we consider here mainly the scenario of offline distillation (noise-image pairs of teacher are generated before training) due to compute resources.
\vspace{-5pt}
\subsection{Design Space Exploration and Principles}
In this section, we look at the impact of various design knobs (mentioned in~\cref{sec: Design description}) on the performance and efficiency of the DiTs. Before sweeping the depth and width knobs, we first identify the best distillation setup based on existing methods and loss functions in~\cref{tab: distillation setup ablation}. The huge impact of using LPIPS loss~\citep{zhang2018unreasonable} is obvious for both GET~\citep{geng2024one} and DMD~\citep{yin2024one}. The simpler training setup of GET provides better results compared to DMD.

\begin{figure}[h]
    \begin{subfigure}{0.32\linewidth}
        \centering
        \includegraphics[width=\linewidth]{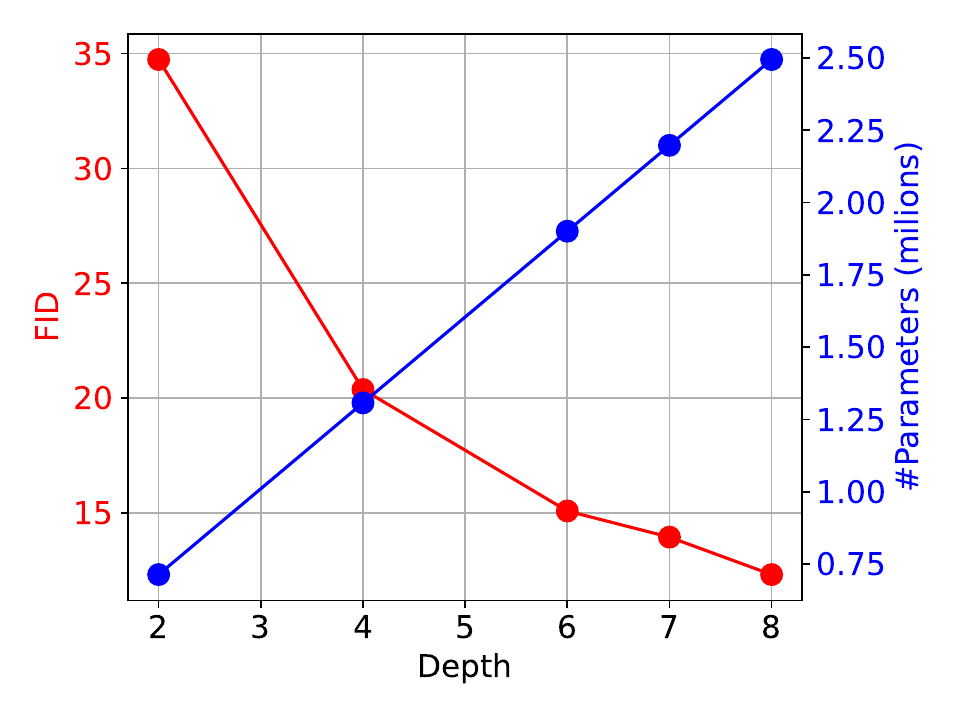}
        \subcaption{\label{fig: depth v/s FID}}
    \end{subfigure}
    \hfill
    \begin{subfigure}{0.32\linewidth}
        \centering
        \includegraphics[width=\linewidth]{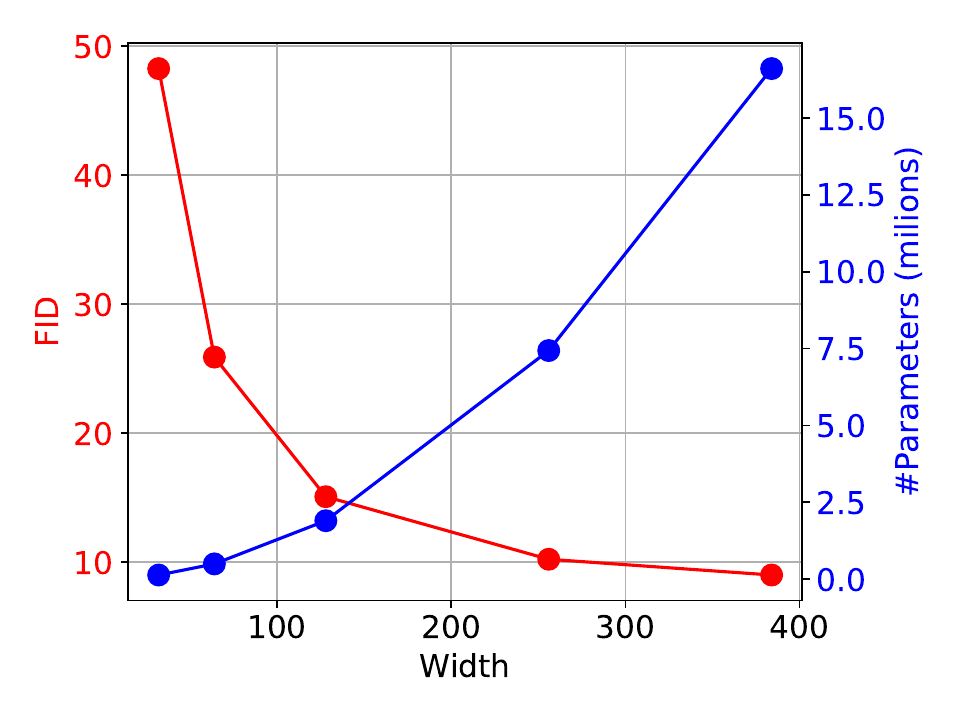}
        \subcaption{\label{fig: width v/s FID}}
    \end{subfigure}
    \hfill
    \begin{subfigure}{0.33\linewidth}
        \centering
        \includegraphics[width=\linewidth]{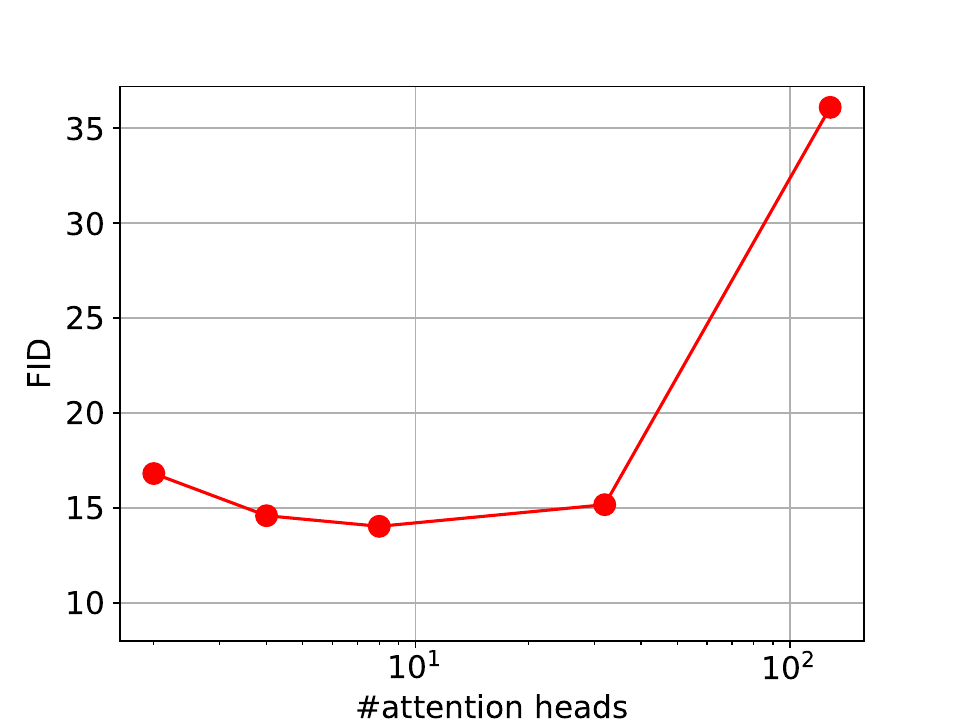}
        \subcaption{\label{fig: heads v/s FID}}
    \end{subfigure}
    \vspace{-5pt}
    \caption{\label{fig: FID variation with depth and width} (a) shows the impact of depth on FID and no. of parameters. It can be observed that the no. of parameters increases linearly with the increase in depth. (b) shows the impact of width on FID and no. of parameters. We see that the no. of parameters increases quadratically with increase in width. Both (a) \& (b) show diminishing returns if increased independently and the performance (FID) does not scale in the same proportion as the increase in parameters, especially for width. (c) impact of no. of attention heads on FID. When the depth ($d$) or width ($w$) or attention heads ($h$) is not changing in any of these plots, it is assumed to be $d=6$, $w=128$ and $h=4$ respectively.}
\end{figure}

\cref{fig: depth v/s FID} \&~\ref{fig: width v/s FID} indicate how the depth $d$ (no. of layers) and width $w$ (embedding dimension) affect the FID and no. of parameters. The key takeaway is that increasing only one quantity without changing the other results in diminishing returns. When it comes to no. of attention heads $h$ (which can only take factors of $w$ as values), we find there is a sweet spot in the middle as shown in~\cref{fig: heads v/s FID}.


\vspace{-5pt}
\begin{minipage}{0.47\columnwidth}
\centering
\captionof{table}{Compares FIDs for different distillation setups and loss functions for DiT. Effect of classifier-free guidance has been excluded for simplicity. The GET setup with LPIPS loss is the most effective.}
\label{tab: distillation setup ablation}
\begin{tabular}{|c|c||c|}
    \hline
    Setup & Loss  & FID   \\ \hline \hline
    GET~\citep{geng2024one}   & L1    & 45.60 \\ \hline
    GET~\citep{geng2024one}   & L2    & 42.73 \\ \hline
    GET~\citep{geng2024one}   & LPIPS & \textbf{18.30} \\ \hline
    DMD~\citep{yin2024one}   & LPIPS & 20.92 \\ \hline
\end{tabular}
\end{minipage}
\hfill
\begin{minipage}{0.5\columnwidth}
\centering
\captionof{table}{Design-space exploration when the parameter size is 0.42M. The latency (L) for generating 50k images is measured on NVIDIA Jetson Nano.}
\label{tab: 0.42M params ablation}
\begin{tabular}{|c|c|c|c|c|c|}
\hline
Name & $d$ & $w$   & $h$ & L (s) & FID   \\ \hline \hline
wider & 1 & 128 & 8 & 97.5 & 57.07\\ \hline
wide & 2 & 96  & 8 & 123.6 & 34.78\\ \hline
lower heads & 5 & 64  & 4 & 162.6 & 29.66\\ \hline
\textbf{proposed} & \textbf{5} & \textbf{64}  & \textbf{8} & 198.7 & 29.14\\ \hline
deep & 9 & 48  & 6 & 259.0 & 28.52\\ \hline
deeper & 21 & 32  & 4 & 411.7 & 32.81\\ \hline
\end{tabular}
\end{minipage}

\textbf{Design Principles}: From the experiments in~\cref{fig: FID variation with depth and width} and~\cref{tab: distillation setup ablation} we provide the below guidelines:
\begin{tcolorbox}[%
    colback=gray!10,    
    colframe=black,     
    rounded corners,
    boxrule=1pt,        
    arc=6pt             
  ]
\begin{enumerate}[noitemsep]
    \item Use LPIPS loss when performing distillation for diffusion tasks.
    \item Choose depth $d \approx \floor{\log_2w}$ with respect to the width $w$ subject to satisfying the model parameter constraint.
    \item Choose number of attention heads $h = \min \text{median}(\{\text{factors}(w)\})$.
\end{enumerate}
\end{tcolorbox}

\begin{figure}[h]
    \begin{subfigure}{0.32\linewidth}
        \centering
        \includegraphics[width=\linewidth]{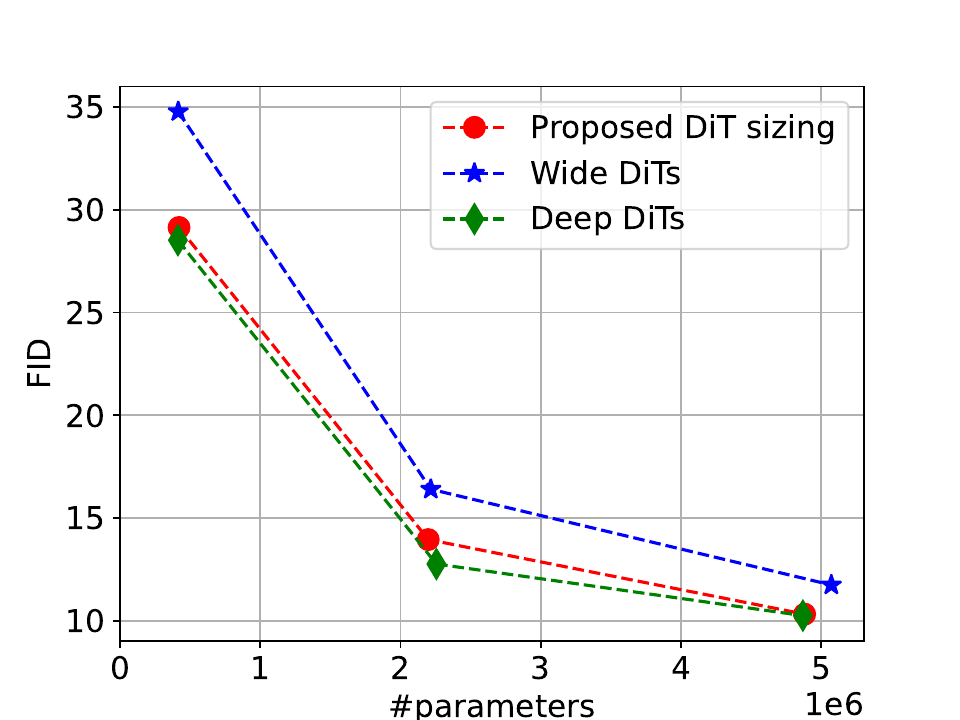}
        \subcaption{\label{fig: parameters v/s FID}}
    \end{subfigure}
    \hfill
    \begin{subfigure}{0.32\linewidth}
        \centering
        \includegraphics[width=\linewidth]{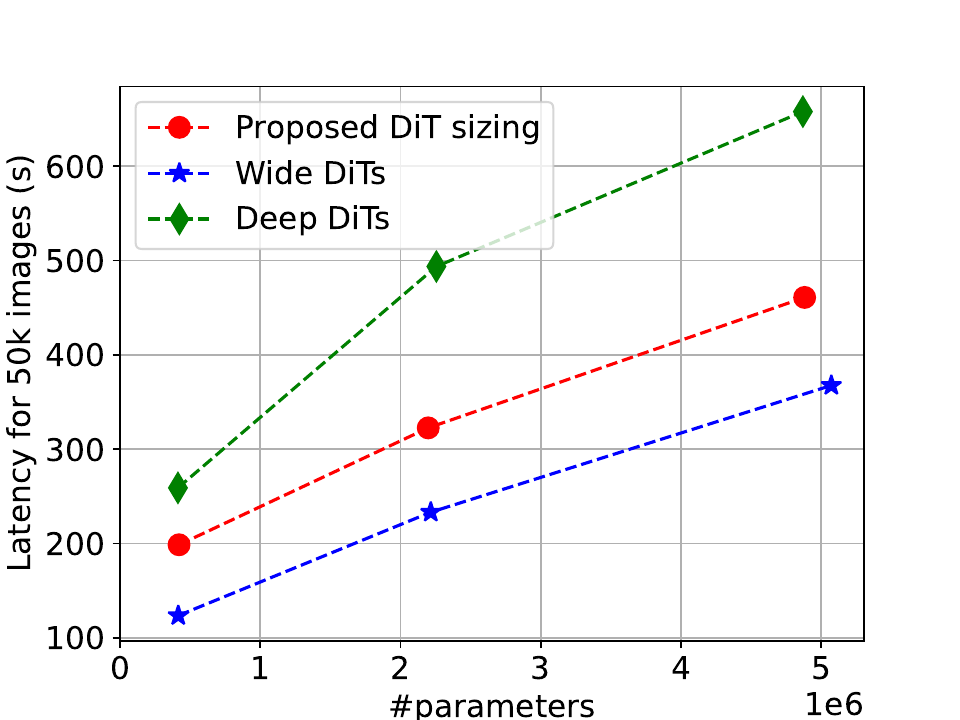}
        \subcaption{\label{fig: parameters v/s latency}}
    \end{subfigure}
    \hfill
    \begin{subfigure}{0.32\linewidth}
        \centering
        \includegraphics[width=\linewidth]{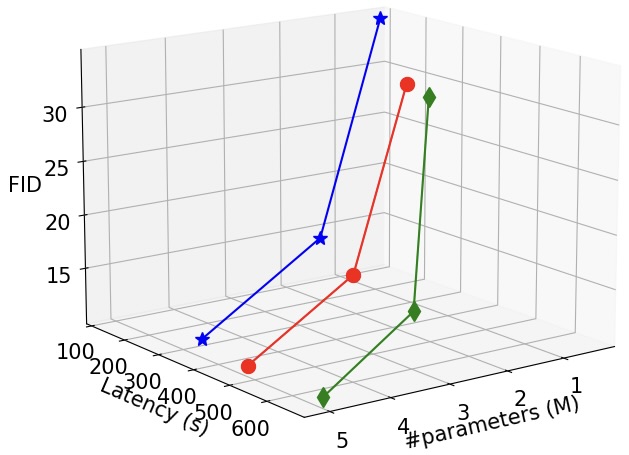}
        \subcaption{\label{fig: parameters v/s FID v/s latency}}
    \end{subfigure}
    \vspace{-5pt}
    \caption{\label{fig: Impact of parameters on FID and latency} (a) Impact of number of parameters on FID when comparing deep DiTs, wide DiTs and the proposed sizing for DiTs. (b) Impact of parameters on the latency (on NVIDIA Jetson Nano). Wide DiTs have an observably worse performance while the deep DiTs have significantly higher latency due to serial processing.~\cref{tab: 0.42M params ablation} shows the FID worsens going any deeper/wider. Our proposed sizing provides almost optimal FID at a much smaller latency. (c) The 3-dimensional trade-off highlighting the \#params (memory) v/s FID (image quality) v/s latency (image frame rate) amidst optimal DiTs.}
    \vspace{-15pt}
\end{figure}

The simulations that validate the design rules only explored widths which were of the form $\{2^n, 3.2^n\}$ since these utilize hardware most effectively. Above, the median for an even set of numbers is taken to be both the middle numbers.~\cref{fig: Impact of parameters on FID and latency} and~\cref{tab: 0.42M params ablation} show the benefits of our design principles (our proposed sizing is in bold) which achieve close to optimal FID for a much lower latency. One interesting aspect in~\cref{tab: 0.42M params ablation} is that when we reduce the attention heads from our proposed suggestion of 8 to 4, the latency reduces. Since this could related to the algorithm-microarchitecture mapping in NVIDIA Jetson Orin Nano, we \textit{do not} include this factor in our design principles.

\vspace{-10pt}
\subsection{Impact of TA \& MI1 methods and SOTA comparison}
\vspace{-5pt}
\cref{tab: TA results} suggests that feature distillation using the TA is not helpful. Only the distillation directly with the TA (architecture details in~\cref{appsubsec: TA}) provides marginal benefits over the baseline obtained from our design principles. MI1 also performs worse compared to baseline as shown in~\cref{tab: MI1 results}. While these results are poor, we include them in support of the recent movement for publishing negative results~\citep{blaas2025can, tafreshi2023proceedings, guo2025deepseek}. A crucial aspect to note is that the constraint on the intermediate layers is not the reason for worse performance since (2, 4, 6) performs better than (3, 6). Lastly, we compare our baseline approach with the only SOTA diffusion-based transformer model~\citep{geng2024one} that does model parameter distillation. We beat them on all metrics - model size, FID and latency (more on SOTA comparison in~\cref{appsubsec: SOTA}).

\begin{minipage}{0.47\columnwidth}\centering
\captionof{table}{Impact of feature distillation using the TA method. The penultimate transformer layers of the TA and student are matched as indicated by~\citep{muralidharan2024compact}. Distilling with only TA~\citep{mirzadeh2020improved} provides the best solution.}
\label{tab: TA results}
\begin{tabular}{|c|c|c|c|}
\hline
Teacher & TA & TA features & FID   \\ \hline \hline
\checkmark &  &  & 14.03\\ \hline
 & \checkmark &  & \textbf{13.99}\\ \hline
 & \checkmark & \checkmark & 14.36\\ \hline
\checkmark & \checkmark & \checkmark & 14.28\\ \hline
\end{tabular}
\end{minipage}
\hfill
\begin{minipage}{0.50\columnwidth}\centering
\captionof{table}{Impact of using MI1.}
\label{tab: MI1 results}
\begin{tabular}{|c|c|c|}
\hline
Mapped Layers & timesteps & FID   \\ \hline \hline
Baseline & - & \textbf{14.03}\\ \hline
(3, 6) & (0.5, 0) & 14.49\\ \hline
(2, 4, 6) & (0.66, 0.33, 0) & 14.32\\ \hline
\end{tabular}

\captionof{table}{SOTA comparison. Params (P) \& Latency (L) for generating 50k images on A100 GPUs.}\centering
\label{tab: SOTA comparison}
\resizebox{0.95\columnwidth}{!}{
\begin{tabular}{|c|c|c|c|}
\hline
Method & P & FID & L (s)  \\ \hline \hline
GET\citep{geng2024one} & 8.6M & 12.93 & 6.46\\ \hline 
GET (Our principles) & 5.3M & \textbf{10.21} & 7.01\\ \hline
DiT-Nano & \textbf{5M} & 10.32 & \textbf{3.66}\\ \hline
\end{tabular}
}
\end{minipage}
 



\vspace{-5pt}
\section{Conclusion}
\vspace{-5pt}
We perform a thorough design space exploration of DiT distillation and provide design principles to obtain SOTA DiTs for a given model size. When these DiTs are implemented on NVIDIA Jetson Orin Nano, we identify a key trade-off between model performance-size-speed which can direct future researchers on practical areas to innovate. We hope the above guidelines serve as a reference when distilling DiTs for larger and more practically-relevant tasks. Through this paper, we also want to emphasize the practice of creating strong and obvious baselines (which was already SOTA in our case) before comparing the novel methods with prior works. Even though the TA method beats our baseline marginally, we conclude that the student model designed from our principles is a better option compared to the TA and MI1 methods due to cheaper training cost. Future directions can include justifying the above guidelines analytically, or expanding the design-space to knobs like MLP ratio and diffusion timesteps, or having custom attention heads for each layer~\citep{michel2019sixteen}, especially since there is an impact on latency with changing attention head size (see~\cref{tab: 0.42M params ablation}).



\clearpage
\bibliography{iclr2025_delta}
\bibliographystyle{iclr2025_delta}
\clearpage
\appendix
\textbf{Acknowledgements}: The author would like to thank Prof. Naresh Shanbhag for his extensive support and discussions during this project. The author would also like to thank Kaining Zhou, Soonha Hwang and Nathan Chiang for the incredible collaboration out of which this work was a part of. Lastly, this work was supported by the Center for the Co-Design of Cognitive Systems (CoCoSys) funded
by the Semiconductor Research Corporation (SRC) and the Defense Advanced Research Projects Agency (DARPA).
\vspace{-5pt}
\section{Related Works}
\vspace{-5pt}
\label{app: related works}
\subsection{Diffusion Models and Architectures}
Diffusion Models were originally introduced by~\citep{sohl2015deep} and had a rapid rise later due to the works of~\citep{song2019generative, ho2020denoising, song2020score}.~\citep{karras2022elucidating} showed comprehensively how convolution-based U-Nets can be adapted to perform diffusion efficiently. More recently, with the advent of transformers, Diffusion Transformers (DiTs)~\citep{peebles2023scalable}, Scalable Interpolant Transformers (SiTs)~\citep{ma2024sit}, Hourglass DiTs (HDiTs)~\citep{crowson2024scalable} and so on have been proposed to show the scalability of the transformer architecture for diffusion. Our work only looks at the distillation of DiTs~\citep{peebles2023scalable} even though the conclusions can be extended to the other transformer architectures mentioned above.

\subsection{Efficient Diffusion Transformers}
Even though many efficiency techniques~\citep{wang2024patch, li2023q, zhao2023mobilediffusion} have been proposed for convolution-based diffusion models, we will focus here on the ones proposed for transformers. Quantization~\citep{chen2024q, wu2024ptq4dit} and pruning~\citep{fang2024tinyfusion} have been proposed for designing parameter-efficient DiTs. Efficient attention mechanisms~\citep{pu2025efficient, yang2025inf} have also been proposed to perform efficient diffusion using transformers. However, all these methods are orthogonal to our approach of parameter-reduction using distillation. Our approach is the first to reduce both parameters and diffusion timesteps using distillation.

\subsection{Timestep Distillation of Diffusion Models}
To improve the slow generation process of diffusion models, extensive research has been done to reduce the timesteps through distillation~\citep{luhman2021knowledge, salimans2022progressive, meng2023distillation, yin2024one, yin2024improved, zhou2024score}. Progressive distillation~\citep{salimans2022progressive} reduces the number of timesteps by two during each distillation stage, thus incurring a large training cost.~\citep{meng2023distillation} propose a classifier-free distillation method to generate images using 1-4 timesteps.~\citep{yin2024one, yin2024improved} use adversarial setups and distribution matching losses to perform one-step generation. Trajectory and consistency-based distillation~\citep{berthelot2023tract, song2023consistency, zheng2024trajectory} has also been considered to improve the speed of diffusion model generation. However, these methods are mainly proposed for U-Nets, do not consider parameter-reduction in models and hence typically initialize the student with the teacher at the beginning of training. Our problem looks at how to reduce both the parameters and timesteps for DiTs through distillation, which is considerably more challenging than just doing the latter.~\citep{geng2024one, teephysics} which do look at transformer distillation with a different student architecture mainly focus on the impact of equilibrium and physics-informed models, respectively, rather than pushing the parameters and compute down.

\subsection{SOTA Comparison}
\label{appsubsec: SOTA}
The only fair comparison for our method is~\citep{geng2024one}. Most works~\citep{yin2024one, yin2024improved, berthelot2023tract} for diffusion distillation have focused solely on timestep distillation and the ones which do consider parameter distillation~\citep{teephysics, dockhorn2023distilling} are for U-Nets and do not focus on providing design principles like us. The methods which use U-Nets for timestep distillation directly initialize their student model with the teacher model before starting training, a trick not available for methods like ours where there is architectural incompatibility in student-teacher networks. Hence we only compare with~\citep{geng2024one} which does parameter distillation for transformer-based diffusion. We also note that an expression similar to ours for depth was obtained in~\citep{wies2021transformer} independently and from a theoretical standpoint for different modality and non-distillation setup. This instills confidence in our design principles.
\section{Motivation and Implementation for Algorithms}
\label{app: algorithms}
\subsection{GET setup}
\textbf{Motivation}: A simple distillation setup that fulfills the essence of knowledge distillation for diffusion.

\textbf{Implementation}: Though this setup is same as~\citep{geng2024one}, we explain it here and in~\cref{app: setup} for completeness. The noise-image pairs are generated offline prior to training and employed as a dataset. This is necessary when training-compute budget is limited since it is more expensive to have an online teacher model generate samples on the fly. The teacher noise is fed as input to the student model and it is matched to the teacher image using a loss function.

\[
\mathcal{L}_{\text{GET}} =  \mathcal{L}_{\text{LPIPS}} (\mathbf{x}, \hat{\mathbf{x}}(\mathbf{z}, c)), \;\;\;\; (\mathbf{z}, c, \mathbf{x})\sim \mathcal{D}_{\text{teach}}
\]
where $\mathbf{z}, c, \mathbf{x}$ represent the noise, class label and image drawn from the offline teacher dataset $\mathcal{D}_{\text{teach}}$, respectively. Here $\hat{\mathbf{x}}(\mathbf{z}, c)$ represents the output from the model which we will represent as $\hat{\mathbf{x}}$ moving forward.

\subsection{TA method}
\label{appsubsec: TA}
\textbf{Motivation}: Since our approach looks at extremely tiny models (0.5M-5M model size) compared to the teacher model (EDM~\citep{karras2022elucidating} with 62M parameters), the teaching assistant model helps bridge this disparity. Apart from that, to overcome the difficulty of feature distillation using a U-Net teacher, we use a DiT model for the TA to explore the benefits of feature distillation. We pick a reasonable choice for the TA architecture (DiT model with $(d, w, h) = (12, 384, 12)$) and leave the exploration of TA network architecture for future work.

\textbf{Implementation}: For the feature matching we take the transformer layer outputs from both TA and the student (expansion tensor is used for the student to match the TA width) and minimized using LPIPS loss. Based on~\citep{muralidharan2024compact} we match the outputs from the penultimate transformer layer. The non-feature TA distillation approach is similar to the regular GET setup.

\[
\mathcal{L}_{\text{GET}} =  \lambda_0\mathcal{L}_{\text{LPIPS}} (\mathbf{x}, \hat{\mathbf{x}}) + \lambda_1\mathcal{L}_{\text{LPIPS}} (\mathbf{x}^{\text{TA}}, \hat{\mathbf{x}}) + \lambda_2\sum_l\mathcal{L}_{\text{LPIPS}} (\mathbf{x}_l^{\text{TA}}, \hat{\mathbf{x}}_l), \;\;\;\; (\mathbf{z}, c, \mathbf{x})\sim \mathcal{D}_{\text{teach}}
\]
 Here the $\mathbf{x}^{\text{TA}}$, $\hat{\mathbf{x}}$, $\mathbf{x}_l^{\text{TA}}$ and $\hat{\mathbf{x}}_l$ represent the image output of the TA model, image output of the student model, intermediate layer $l$ output of the TA model and intermediate layer $l$ output of the student model, respectively. Depending on the configuration in~\cref{tab: TA results}, we make $\lambda_0\neq0$ (row 1, 4) or $\lambda_1\neq0$ (row 2, 3, 4) or $\lambda_2\neq0$ (row 3, 4) or all are non-zero (row 4). 

\subsection{MI1 method}
\label{appsubsec: MI1}
\textbf{Motivation}: Even though many methods try to perform one-step diffusion distillation~\citep{yin2024one, yin2024improved}, they do not explicitly map the intermediate outputs to each layer of the student. We do so since smaller models, unlike bigger models, have limited capacity to figure out how to effectively generate images. An explicit diffusion-based layer-wise guidance should ideally help teach them how to generate images using a "diffusion trajectory" instead of having to figure out how to do it.

\textbf{Implementation}: The goal of this method is to explicitly teach the small student model how to generate images by modeling the multi-step diffusion process in a single forward pass. We achieve this by mapping certain diffusion timestep samples to specific layers of the DiT. For example, if you consider a 1000 step diffusion process of the teacher(from a reverse diffusion perspective 0th step is the noise sample and 1000th step is the image) to be performed in a single pass of a 8-layer student DiT, we map the 250th step to the 2nd layer, 500th to the 4th, 750th to the 6th and the 1000th to the last layer. For a practically feasible distillation, we look at the 18-step EDM teacher~\citep{karras2022elucidating} to be mapped onto a single step of our DiT-Nano. We use the same noise-image pairs generated offline from the GET setup as our starting point. Instead of generating the intermediate timestep samples using the online EDM model, we use the forward probability flow ODE to generate them. We derive below the equations for forward diffusion of the probability flow ODE.

\[
\mathbf{x}(t) = \sqrt{\alpha(t)}\,\mathbf{x}_0 \;+\; \sqrt{1 - \alpha(t)}\,\epsilon,\;\;\;\; \alpha(t) = \frac{1}{\,1 + \sigma(t)^2\,}
\]

\[
\mathbf{x}(t)
= \frac{1}{\sqrt{1 + \sigma(t)^2}}\;\mathbf{x}_0
\;+\;
\frac{\sigma(t)}{\sqrt{1 + \sigma(t)^2}}\;\epsilon
\]
\[
\epsilon
= \frac{
  \mathbf{z} \;-\;\sqrt{\alpha(1)}\,\mathbf{x}_0
}{
  \sqrt{1 - \alpha(1)}
}
\]

\[
\sigma(t) = ({\sigma_{\text{max}}}^{\frac{1}{\rho}} + t({\sigma_{\text{min}}}^{\frac{1}{\rho}}-{\sigma_{\text{max}}}^{\frac{1}{\rho}}))^\rho, \;\;\;\; t=[0,1],\; \sigma_{\text{max}} = 80,\;\sigma_{\text{min}} = 0.02
\]
Here the values of $\sigma(t)$ is taken directly from the implementation of EDM~\citep{karras2022elucidating} while the other parts of the forward probability flow ODE has been derived. In the example shown in~\cref{fig: distillation methodology} and~\cref{tab: MI1 results}, we used $t=\{0.5,0\}$ and $t=\{0.66, 0.33, 0\}$. In the former case, we map the timestep 0.5 to 3rd layer and 0 to last layer of a 6-layer DiT-Nano. In the latter case, we map 0.66 to 2nd layer, 0.33 to the 4th layer and 0 to the last layer of the same model. When we say mapping, we minimize the LPIPS loss between the two objects ($\mathbf{x}(t), \hat{\mathbf{x}}_l$) being mapped. 

\[
\mathcal{L}_{\text{MI1}} = \sum_{l\in \Lambda, t\in \mathcal{T}} \mathcal{L}_{\text{LPIPS}} (\mathbf{x}(t), \hat{\mathbf{x}}_l)
\]

where $\Lambda, \mathcal{T}$ are the sets of layers and timesteps that are being mapped, respectively. The preliminary results which we have reported in~\cref{tab: MI1 results} have been poor and worse than the baseline generated from our guidelines. 
An easy argument or explanation for this could be that MI1 is forcing (constraining) the model to generate images through a diffusion trajectory, whereas it is possibly better for the model to figure the optimal way to generate the image from the input noise on its own. However, if one looks at $t=\{0.5,0\}$ and $t=\{0.66, 0.33, 0\}$, we find the latter to have a lower FID even though it is more constraining, thus countering the argument provided above.
\vspace{-10pt}

\section{Setup}
\label{app: setup}
\vspace{-10pt}
The training or distillation setup closely follows the GET setup~\citep{geng2024one} where as the model setup follows the DiT setup~\citep{peebles2023scalable}. 

\textbf{Data}: For training data we use the noise-image pairs generated from EDM~\citep{karras2022elucidating} VP conditional model provided in the GitHub repository of~\citep{geng2024one}. We only consider such an offline distillation method due to computational constraints of having an online teacher. All our results have been provided for only conditional generation similar to the DiT paper~\citep{peebles2023scalable}. We acknowledge that our results and guidelines are on a small dataset and do not sweep extensively number of design points. However, we only provide our results on CIFAR-10 data due to computational constraints and hope our design principles will enable efficient design of DiTs for larger and more practical datasets and models. 

\textbf{Optimizer}: We use AdamW optimizer with weight decay 0.01, a fixed learning rate of 0.0001, and a global batch size of 256. All experiments have been run for 100 epochs unless stated otherwise. 

\textbf{Models}: Our backbone code for the model is adopted from the DiT paper~\citep{peebles2023scalable} with modifications to accommodate the TA and MI1 methods. We specifically look at DiTs since they are becoming the mainstream models for practical deployment over U-Nets. We also do not consider variants of DiTs~\citep{ma2024sit, crowson2024scalable} since our results can be extended to those. Patch-size of 2 is used for all models.

\textbf{Hardware}: Apart from~\cref{tab: distillation setup ablation} (which was implemented on NVIDIA Quadro RTX 6000), all the training experiments were run on $4\times$ A100 GPUs. While some of the inference has also been on $4\times$ A100 GPUs, the latency calculation for some models has been implemented on NVIDIA Jetson Orin Nano.

\textbf{Inference}: During inference we use an Exponential Moving Average (EMA) model trained with an EMA decay of 0.9999. We also employ classifier-free guidance during inference with a cfg-scale of 1.5. The latency was calculated for 50k images with a batch size 128. All our distillation methods have looked at only performing one-step diffusion with the student.

We will make the code public if accepted.
\section{Extensive Results}
\label{app: results}
\cref{tab: all params ablation} shows the comprehensive results from~\cref{fig: Impact of parameters on FID and latency} in a tabular format. All the results in the main paper and in~\cref{tab: all params ablation} are run for 100 epochs, except the last row in~\cref{tab: all params ablation} which is run for 200 epochs, and GET (Our principles) in~\cref{tab: SOTA comparison} which was run for only 38 epochs! (due to compute constraints, GET training with LPIPS loss is very expensive). IS in~\cref{tab: all params ablation} refers to Inception Score.
\begin{table}[]
\centering
\caption{Design-space exploration for all parameter sizes. The latency for generating 50k images is measured on NVIDIA Jetson Orin.}
\label{tab: all params ablation}
\begin{tabular}{|c|c|c|c|c|c|c|}
\hline
\#Params & D & W   & H & Latency (s) & FID & IS  \\ \hline \hline
\multirow{6}{*}{0.42M}& 1 & 128 & 8 & 97.5 & 57.07 & 6.38\\ 
& 2 & 96  & 8 & 123.6 & 34.78 & 7.48\\ 
& 5 & 64  & 4 & 162.6 & 29.66 & 7.76\\ 
& \textbf{5} & \textbf{64}  & \textbf{8} & 198.7 & 29.14 & 7.79\\ 
& 9 & 48  & 6 & 259.0 & 28.52 & 7.78\\ 
& 21 & 32  & 4 & 411.7 & 32.81 & 7.41\\ \hline 
\multirow{3}{*}{2.2M}& 3 & 192  & 12 & 233.2 & 16.40 & 8.78\\ 
& \textbf{7} & \textbf{128}  & \textbf{8} & 322.6 & 13.95 & 8.90\\ 
& 13 & 96  & 8 & 493.7 & 12.76 & 8.92\\ \hline
\multirow{2}{*}{5M}& 4 & 256  & 16 & 367.6 & 11.74 & 9.04\\ 
& 16 & 128  & 8 & 657.9 & 10.25 & 9.15\\ 
& \textbf{7} & \textbf{192}  & \textbf{12} & 460.9 & 10.32 & 9.16\\ 
5M (200 epochs)& \textbf{7 }& \textbf{192}  & \textbf{12} & 458.6 & \textbf{8.90} & \textbf{9.31}\\ \hline 
\end{tabular}
\end{table}

\end{document}